%% file: main_matter.tex
\documentclass[letterpaper, 10 pt, journal, twoside]{ieeeconf}
\usepackage{amsmath,amssymb,amsfonts}
\usepackage{algorithmic}
\usepackage{graphicx}
\graphicspath{ {./media/} }
\usepackage{textcomp}
\usepackage[table]{xcolor}
\let\labelindent\relax
\usepackage{enumitem}
\usepackage[hidelinks]{hyperref}
\usepackage[acronym]{glossaries}
\usepackage{multirow}
\usepackage[per-mode = symbol]{siunitx}
\usepackage{booktabs}
\usepackage{todonotes}
\usepackage{tablefootnote}
\usepackage{import}
\usepackage[backend=biber,style=numeric-comp,sorting=none]{biblatex}
\usepackage{tikz, pgfplots}
\pgfplotsset{compat=1.18} 
\usetikzlibrary{calc, angles,quotes,arrows.meta,decorations.markings}
\definecolor{edits}{rgb}{0.0, 0.0, 0.0}
\definecolor{light-gray}{gray}{0.9}
\definecolor{delftblue}{RGB}{0, 166, 214}
\newcolumntype{C}[1]{>{\centering\arraybackslash}m{#1}}

\newacronym{tms}{TMS}{Tether Management System}
\newacronym{uav}{UAV}{Unmanned Aerial Vehicle}
\newacronym{ti}{TI}{Tether-Inertial}
\newacronym{tuav}{TUAV}{Tethered Unmanned Aerial Vehicle}
\newacronym{esa}{ESA}{European Space Agency}
\newacronym{nasa}{NASA}{National Aeronautics and Space Administration}
\newacronym{adc}{ADC}{Analog-to-Digital Converter}
\newacronym{imu}{IMU}{Inertial Measurement Unit}
\newacronym{ros}{ROS}{Robot Operating System}
\newacronym{mocap}{MoCap}{Motion Capture}
\newacronym{pd}{PD}{Proportional-Derivative}
\newacronym{cad}{CAD}{Computer-Aided Design}
\newacronym{estec}{ESTEC}{European Space Research and Technology Centre}
\newacronym{gp}{GP}{Gaussian Process}
\newacronym{rmse}{RMSE}{Root Mean Square Error}
\newacronym{prl}{PRL}{\protect\prlname}
\newacronym{ekf}{EKF}{Extended Kalman Filter}

\def\BibTeX{{\rm B\kern-.05em{\sc i\kern-.025em b}\kern-.08em
    T\kern-.1667em\lower.7ex\hbox{E}\kern-.125emX}}
\begin{document}

\linespread{0.985}
\setlist{nosep}

\newcommand{\prlname}{\ifdefined\anon\textit{[anonymous lab] }\else Planetary Robotics Lab \fi}
\newcommand{\estecname}{\ifdefined\anon\textit{[anonymous location] }\else European Space Agency's ESTEC \fi}

\title{\vspace{0.5cm}Tether-Inertial Localization for Planetary Drones}

\markboth{IEEE Robotics and Automation Letters. Preprint Version. Accepted July, 2026}
{Van Loon \MakeLowercase{\textit{et al.}}: Tether-Inertial Localization for Planetary Drones} 

\author{
    \ifdefined\anon
        Author 1$^{1}$, Author 2$^{2}$, Author 3$^{3}$, Author 4$^{3}$ and Author 5$^{2}$
    \else 
        Dielof van Loon$^{1}$, Anton Bredenbeck$^{2}$, Lennart Puck$^{3}$, Martin Azkarate$^{3}$ and Salua Hamaza$^{2}$
    \fi
    \thanks{
        Manuscript received: March, 6, 2026; Revised June, 13, 2026; Accepted July, 28, 2026.
    }
    \thanks{
        This paper was recommended for publication by Editor Giuseppe Loianno upon evaluation of the Associate Editor and Reviewers comments.
    }
    \thanks{
        \ifdefined\anon
            $^{1}$Affiliation 1.
        \else
            $^{1}$Dept. of Cognitive Robotics, Faculty of Mechanical Engineering, TU Delft, The Netherlands.
        \fi
    }
    \thanks{
        \ifdefined\anon
            $^{2}$Affiliation 2.
        \else
            $^{2}$Biomorphic Intelligence Laboratory, Dept. of Control \& Operations, Faculty of Aerospace Engineering, TU Delft, The Netherlands.
        \fi
    }
    \thanks{
        \ifdefined\anon
            $^{3}$Affiliation 3.
        \else
            $^{3}$Planetary Robotics Lab, European Space Agency, Noordwijk, The Netherlands.
        \fi
    }
    \thanks{
        Digital Object Identifier (DOI): see top of this page.
    }
}



\maketitle

\begin{abstract}
Recent developments in planetary exploration have shown the potential of \glspl{uav}, such as the Ingenuity helicopter that provided valuable mapping data. However, limited payload capabilities constrain the flight times and compute available for localization which restrict their applicability. By providing a tethered connection, issues such as battery and computational constraints are offloaded to the base rover. At the same time, the cable can be exploited for non-drifting localization. This work presents a novel Tether-Inertial Localization approach that uses tether length, and angle measurements to estimate the \gls{uav} {\color{edits} position} relative to its base. The method combines a computationally efficient analytical catenary model with a \gls{gp} residual error compensation. This accounts for systematic sensor inaccuracies and model limitations.
Experimental validation across circular, triangular, and figure-eight trajectories with tether lengths up to \SI{4.5}{\meter} and a total flight time of 37 minutes demonstrates the effectiveness of the proposed approach. Using only tether-based position estimates for feedback, the analytical catenary model achieves an average RMSE of \SI{7.4}{\centi\meter}, which is further reduced to \SI{5.2}{\centi\meter} through \gls{gp}-based residual compensation, {\color{edits} one order of magnitude better than the state-of-the-art}. These results establish Tether-Inertial Localization as a practical alternative to vision- and GNSS-based localization for {\color{edits}\glspl{tuav}}.
\end{abstract}


\vspace{-0.1cm}\section{Introduction}

\PARstart{T}{he} growing focus on celestial body exploration for applications including resource extraction and habitability assessment has increased the need for autonomous robots for planetary exploration. Current examples include the Perseverance rover~\cite{mangold2021perseverance} and the Ingenuity helicopter~\cite{balaram2021ingenuity}.
Aerial robots can significantly enhance planetary exploration missions by enabling more reliable detection of distant objects, extending communication range through radio beacons beyond the horizon, and collecting data from a mobile, elevated sensing platform.
The Ingenuity helicopter proved the viability of aerial planetary robotics on Mars throughout its 72 missions, providing valuable mapping data {\color{edits} from flights at a nominal altitude of \SI{5}{\meter}}; nevertheless, its final mission concluded in a crash caused by insufficient features for its vision-based localization system~\cite{NASA_2024}.
The Ingenuity experience demonstrates that \glspl{uav} still face substantial challenges, primarily due to restricted payload capabilities. This results in limited flight duration due to battery constraints and limited on-board processing of image data for localization, stemming from strict mass and power budgets.
One way to address these challenges is to incorporate a tether that links the \gls{uav} to a companion rover. The tether can provide continuous power, offboard computation, and reliable communication links, thus extending flight duration and enabling more computation-demanding tasks. While \gls{tuav} systems are commercially accessible~\cite{Folorunsho2024-bj}, a significant gap remains: none of these systems exploit the physical properties of the tether as a source for localizing the \gls{uav} relative to the rover.
\begin{figure}
    \centering
    \ifdefined\anon
    \includegraphics[width=\columnwidth, trim={0 2.0cm 0 0}, clip]{media/platform_rover2_anonymous.jpg}
    \else
    \includegraphics[width=\columnwidth, trim={0 5.0cm 0 0}, clip]{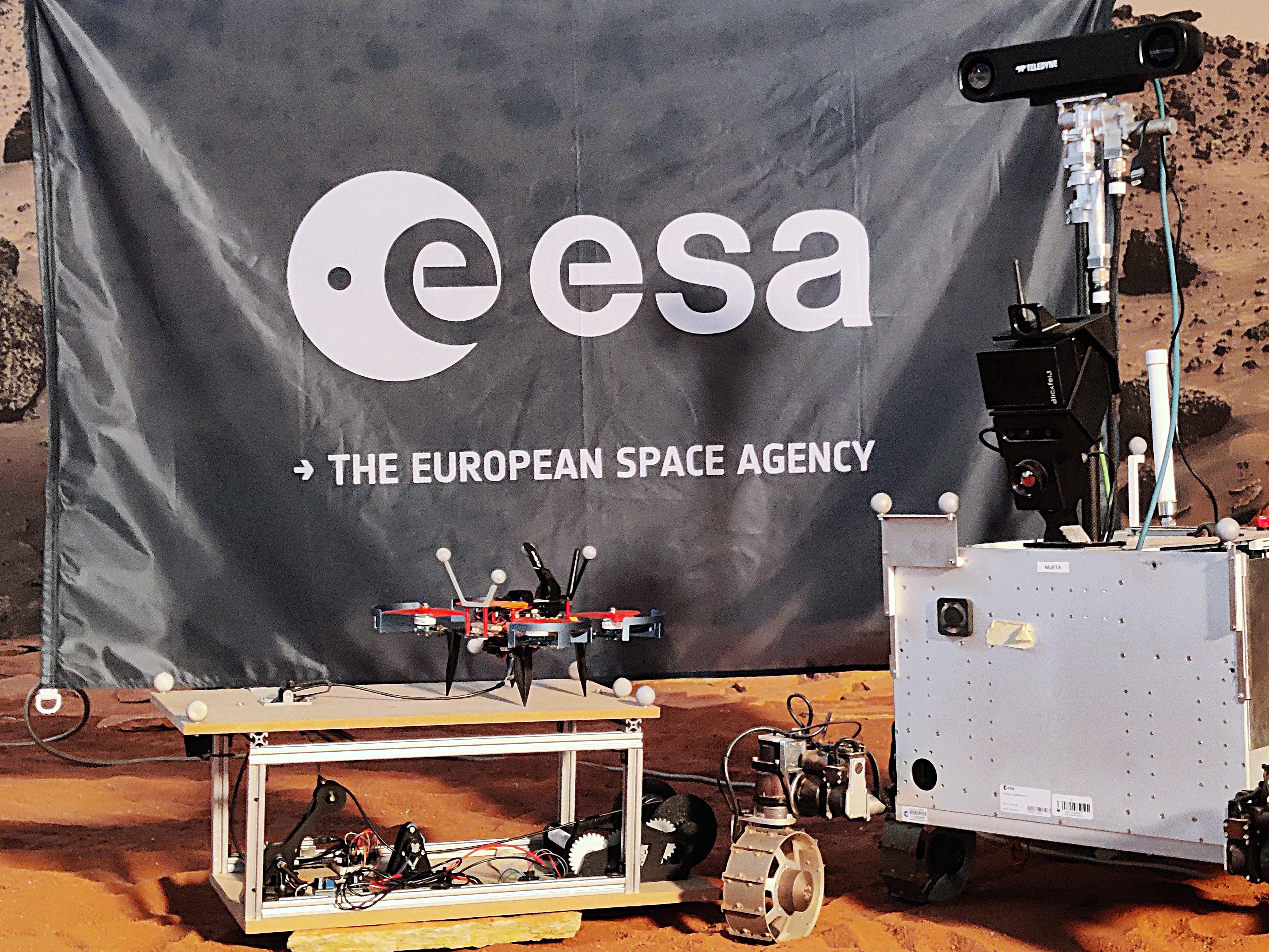}
    \fi
    \vspace{-0.6cm}
    \caption{Our custom-made \acrlong{tms} with the QAV250 drone next to the rover in the Mars yard of the \prlname at the \estecname.
    The drone uses the tether-inertial localization to compute its position based on the properties of the tether.}
    \label{fig:esa_pic}
    \vspace{-0.5cm}
\end{figure}
In this work, we introduce a novel {\color{edits}tether-inertial} Localization approach using a catenary model and a \gls{gp}. Our system incorporates length, angle, and tension sensors to measure all relevant tether state variables, enabling accurate and non-drifting estimation of the \glspl{uav} position {\color{edits} relative to the rover}.
Thereby enabling robust position control even in feature-sparse environments.
The contributions of this work are:
\begin{itemize}
\item[-] A Tether-Inertial localization approach that provides a non-drifting position estimate and its {\color{edits} analytically computed uncertainty} from tether measurements.
\item[-] {\color{edits}A variance-gated \gls{gp} trained on pairs of tether states and residual errors to compensate localization errors while preserving the analytical accuracy guarantees.}
\item[-] The comprehensive experimental validation of the {\color{edits}tether-inertial} Localization approach using a custom-built \gls{tms}.
\end{itemize}\newpage

\noindent{\color{edits} To the best of our knowledge, this is the first work to demonstrate closed-loop tether-based localization of a \gls{tuav} with sub-\SI{10}{\centi\meter} accuracy and variable tether lengths, enabling precise positioning over a large volume using only the tether-based estimate.}

\section{Related Work}

\glspl{tuav} have been deployed across applications requiring continuous power supply from the ground and a reliable communication link~\cite{boukoberine2019power}, for surveillance~\cite{motlagh2017uav}, infrastructure inspection~\cite{10109426, oxpecker}, and defense operations~\cite{tethered_defense}.
Beyond power and communication, the physical tether connection also enables controlled landing assistance~\cite{oxpecker, lima_vectors} and provides operational constraints for flying in restricted airspace~\cite{drones_review}.

Several approaches have been explored for localizing \glspl{tuav} in GNSS-denied environments~\cite{schuster2024tactile}.
Methods that explicitly use tether information include:
(i) approaches that assume the tether remains taut~\cite{wo_cable_sensor, tognon}, and
(ii) approaches that account for the tether's flexible nature~\cite{lima_multi, indoor_uav, ugv_uav}.
The work presented in~\cite{indoor_uav} employs a commercial \gls{tms} that measures elevation and azimuth angles of the tether departure point.
The tether angle at the drone is inferred from the drone's attitude relative to the horizontal plane.
The position estimation uses a taut tether assumption with corrections for slacking tether effects. Indoor static validation with motion capture ground truth achieved \SI{0.37}{\meter} average error at maximum tether lengths of \SI{3.25}{\meter}. Borgese et al.~\cite{ugv_uav} use custom hardware mounted on a rover which measures angles at both ends of the tether, but uses a fixed length tether. The tether position model uses catenary equations, making the position estimate account for the curvature of the tether. The method is tested with a few static points and two outdoor flight tests, using RTK-GPS as ground truth and position feedback. Here, the authors {\color{edits} report accuracies of \SI{0.73}{\meter} and \SI{0.60}{\meter} for two different outdoor waypoint trajectories with a fixed length tether yielding an average error of \SI{0.66}{\meter} in flight,} with a \SI{5}{\meter} tether.
Most recently, Lima et al.~\cite{lima_multi} propose a classifier that chooses from three different models: a taut tether, a catenary model<, and a neural network model according to the current state variables.
Using custom-built sensors for estimating length, tension, and angle, the tether state variables are sensed.
The \gls{rmse} from the catenary model is about \SI{1.27}{\meter} and the complete framework reports a \gls{rmse} of \SI{1.1}{\meter}.
In the works~\cite{rico2021trajectory, rico2023analytics}, Rico et al. propose to use a catenary curve to localize sensors along a fixed-length tether connected to a \gls{tuav}.
The authors exploit fixed \glspl{imu} along the tether to estimate the attitude, which is unique along the catenary curve.
This then yields the position of the sensor.
In their work, the authors are able to achieve an \gls{rmse} of \SI{0.95}{\meter} with a \SI{4}{\meter} tether.

Despite these advances, existing works have not {\color{edits} demonstrated localization accuracies below \SI{30}{\centi\meter}, making them insufficient for precise position control~\cite{lima_multi,indoor_uav, ugv_uav, rico2021trajectory, rico2023analytics}}.
Our work addresses this gap with an analytical catenary-based model, augmented by a \glspl{gp} residual correction, achieving accuracies at the \SI{5}{\centi\meter} range.
This enables real-time localization and positioning, an order of magnitude more accurate than {\color{edits} the state-of-the-art}.
We provide an analytical expression for the uncertainty as a function of sensor accuracies and tether state variables, and thoroughly validate our method in real experiments with variable tether lengths of up to \SI{4.5}{\meter}.

\vspace{-0.1cm}\section{Tether-based Localization}

The physical connection between the ground base and the drone imposes a geometric constraint on the drone’s position: it can only occupy locations that are consistent with the current tether configuration.
A constraint that can be exploited to localize the \gls{tuav}.
We model the tether as an idealized hanging chain, assuming a catenary shape, first studied by Leibniz, Huygens, and Bernoulli~\cite{euler1980rational}.
The resulting model has four degrees of freedom and therefore requires four independent measurements for unique identification.
In this work, we use the tether length, the base elevation angle, the base azimuth angle, and the drone elevation angle. 
Furthermore, the catenary model relies on three main assumptions~\cite{Lockwood2007-wn}:
\begin{enumerate}[label=\roman*.]
    \item The tether has negligible bending stiffness w.r.t. its weight {\color{edits} and negligible extensibility at the tension setpoint w.r.t. its overall length}. 
    \item The tether has a uniform mass distribution. 
    \item The tether hangs between exactly two points in a uniform gravitational field.
\end{enumerate}

Note that these assumptions naturally extend to the case of lower gravity environments, as long as the gravitational field is uniform.
However, stiffness of the tether, friction within the system, sensor noise, and other non-linear effects introduce discrepancies to this model.
To capture these effects, we propose a data-driven approach. Using a dataset of ground truth positions and our catenary-based localization, we train a Sparse Gaussian Process~\cite{Rasmussen2005-ep} to predict the residual error.

\subsection{Catenary Model}

\begin{figure}
    \centering
    \input{tikz/catenary}
    \vspace{-0.75cm}
    \caption{3D visualization of the catenary problem showing state variables $\beta_o$, $\beta_d$, $\alpha_o$, and $L$: elevation angles at origin and drone, azimuth angle at origin, the azimuth angle around the z-axis, and tether length, respectively.}
    \label{fig:catenary_2d}
    \vspace{-0.5cm}
\end{figure}
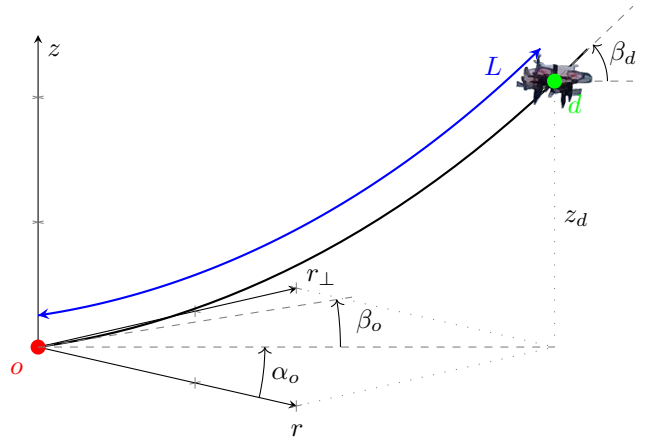

For readability and without loss of generality, we consider the problem as planar. Within this plane, we express the catenary curve as the following function that relates the radial distance (horizontal) with the Z-axis (upwards):
\begin{align}
    z(r)=a\cosh\left(\frac{r-b}{a}\right)+c \label{eq:catfunc}
\end{align}
where $a$, $b$ and $c$ define the curvature and position of the catenary curve~\cite{euler1980rational,dantonio2021catenary}.
At each timestep, the unknown catenary parameters and the drone's radial distance $r_d$ (corresponding to the measured tether length) are determined from the sensor measurements.
Using the base angle $\beta_o$, endpoint angle $\beta_d$, and curve length $L$ shown in~\autoref{fig:catenary_2d}, we establish the following constraints:
\begin{align}
    z(0) & =a\cosh\left(\frac{0-b}{a}\right)+c = 0 \label{eq:constraint1}  \\ 
    \frac{dz}{dr}\bigg|_{r=0} &=\sinh\left(\frac{0-b}{a}\right)=\tan\beta_o \label{eq:constraint2} \\ 
    \frac{dz}{dr}\bigg|_{r=r_d} &=\sinh\left(\frac{r_d-b}{a}\right)=\tan\beta_d \label{eq:constraint3} \\ 
    L &=\int_{0}^{r_d} \sqrt {1+(z'(r))^2} dr \label{eq:constraint4}
\end{align}
Constraint~\eqref{eq:constraint1} fixes the curve to pass through the origin. Constraints~\eqref{eq:constraint2} and~\eqref{eq:constraint3} ensure the curve slope matches the measured elevation angles at the origin and drone positions. Constraint~\eqref{eq:constraint4} enforces that the arc length equals the measured tether length. Taking the derivative of~\eqref{eq:catfunc}, substituting into~\eqref{eq:constraint4}, and integrating from $0$ to $r_d$ yields:
\begin{align}
    L= a [\sinh(\tfrac{r_d-b}{a})-\sinh(\tfrac{0-b}{a})] 
    = a(\tan \beta_d - \tan \beta_o) \label{eq:length-a}
\end{align}
This provides $a$ in terms of known variables. We then solve for the remaining parameters, ultimately giving us $(r,z_d)$, the coordinates of the drone in the projected 2D plane:
\begin{align}
    a &= \frac{L}{\tan \beta_d - \tan \beta_o} \qquad
    b = -a\sinh^{-1}(\tan\beta_o) \label{eq:anb}\\
    c &= -a\cosh\left(\frac{b}{a}\right) \qquad 
    r_d = b+ a\sinh^{-1}(\tan\beta_d) \label{eq:cnx}
\end{align}
By rotating the planar solution by the azimuth angle $\alpha_o$, we finally obtain the full 3D position of the drone:
\begin{align}
    x_d &= r_d\cos{\alpha_o} \qquad
    y_d = r_d\sin{\alpha_o} \\
    z_d & =a\cosh\left(\frac{r_d-b}{a}\right)+c \label{eq:y}
\end{align}
The closed-form solution enables real-time computation with minimal computational overhead, suitable for planetary exploration applications.

\subsection{Model Uncertainty}\label{chap:uncertainty}
As the catenary algorithm is a closed-form solution, the accuracy will depend on the measurement uncertainty of the variables that feed into the algorithm. {\color{edits} By propagating the sensor noise through our model, we can predict the impact of this error on the aleatoric uncertainty of the position measurement.
While this doesn't consider epistemic uncertainty, brought on through unmodeled effects, calibration errors and other biases, this enables analysis of how sensor noise affects the localization accuracy}.
The sensors providing the azimuth and elevation angles are analog sensors, see~\autoref{chap:hw}. This means the \gls{adc}, the reference voltage, and sensor imperfections introduce noise into the system. We determine the level of interference by measuring the standard deviations of the signals during static operation and deriving the partial derivatives w.r.t. each input variable to predict the impact of the sensor noise on a tether-localized position. Using some common definitions to simplify the expressions,`'
{\color{edits}
\begin{align}
 \Delta &= \tan\beta_d - \tan\beta_o, \qquad\
 \Delta_s = s_d - s_o \;\;\text{ with }\\
 s_o &= \sinh^{-1}(\tan\beta_o), \qquad  
 s_d = \sinh^{-1}(\tan\beta_d) \\
 A_o &= \frac{\Delta}{|\cos\beta_o|} - \frac{\Delta_s}{\cos^2\beta_o}, \quad
 A_d = \frac{\Delta}{|\cos\beta_d|} + \frac{s_o - s_d}{\cos^2\beta_d} \\
 B &= \frac{1}{|\cos\beta_d|} - \frac{1}{|\cos\beta_o|}
\end{align}
}
we yield the partial derivatives of $x_d,\,y_d$:

\begin{align}\label{eq:uncertainty}
\frac{\partial x_d}{\partial L} &= \frac{\Delta_s \cos\alpha_o}{\Delta} & & \frac{\partial y_d}{\partial L} = \frac{\Delta_s \sin\alpha_o}{\Delta} \\ 
\frac{\partial x_d}{\partial \beta_o} &= -\frac{L A_o \cos\alpha_o}{\Delta^2} & & \frac{\partial y_d}{\partial \beta_o} = -\frac{L A_o \sin\alpha_o}{\Delta^2} \\
\frac{\partial x_d}{\partial \beta_d} &= \frac{L A_d \cos\alpha_o}{\Delta^2} & & \frac{\partial y_d}{\partial \beta_d} = \frac{L A_d \sin\alpha_o}{\Delta^2} \\
\frac{\partial x_d}{\partial \alpha_o} &= -\frac{L \Delta_s \sin\alpha_o}{\Delta} & & \frac{\partial y_d}{\partial \alpha_o} = \frac{L \Delta_s \cos\alpha_o}{\Delta}
\end{align}
and for $z_d$:
\begin{align}
\frac{\partial z_d}{\partial L} &= \frac{B}{\Delta}\qquad
\frac{\partial z_d}{\partial \alpha_o} = 0\\
\frac{\partial z_d}{\partial \beta_o} &= \frac{L\left(-(\tan\beta_d\tan\beta_o + 1)|\cos\beta_o| + \frac{1}{|\cos\beta_d|}\right)}{\Delta^2 \cos^2\beta_o} \\
\frac{\partial z_d}{\partial \beta_d} &= \frac{L\left(-(\tan\beta_d\tan\beta_o + 1)|\cos\beta_d| + \frac{1}{|\cos\beta_o|}\right)}{\Delta^2 \cos^2\beta_d} \label{eq:uncertaintyend}
\end{align}
Intuitively, these show that within close proximity of the base, the length measurement has a larger influence on the uncertainty than the angle measurement, whereas at large tether lengths, the angle measurements contribute more to the estimate's uncertainty.
{\color{edits} Note that these uncertainty estimates are ill-conditioned in the singular configuration $\Delta \approx 0$, which by catenary geometry implies $z_d \approx 0$. 
In practice this is easily avoided, as the drone flies above the ground base.}

\subsection{GP-Based Residual Error Compensation}
We collect ground truth positions via \gls{mocap} alongside data from the tether sensors and use this data to train a \gls{gp}~\cite{Rasmussen2005-ep} to estimate {\color{edits}the residual error between the tether-localized position and the ground truth position to compensate for unmodeled effects.}
Using these, we can form the input vector which consists of $\begin{bmatrix}x_d, & y_d, & z_d, & \alpha_o, & \beta_o, & \beta_d, & L\end{bmatrix}^\top$.
The 3-dimensional output vector is defined as:
\begin{align}
    \mathbf{e}_d = \begin{bmatrix}x_{gt}-x_d, & y_{gt}- y_d, & z_{gt} - z_d\end{bmatrix}^\top
\end{align}
{\color{edits} This describes a purely geometric relationship, making the residual estimate transferable across different gravitational regimes, similar to the analytical model.}
For training the \gls{gp}, we collect 150k data points from 30 flights with \gls{mocap} in the loop as position feedback to the drone.
Of those 30 test flights, 25 with diverse trajectories (circular, triangular, and figure-eight) are used for training and five for testing.
It is then necessary to sparsify the training data due to poor scalability of \glspl{gp}~\cite{mcintire2016sparse}.
This is done using K-means clustering~\cite{kmeans}, and the Adam optimizer~\cite{adam} and the Evidence lower bound~\cite{elbo} are used for training the \gls{gp}. 
To evaluate the best hyperparameters, we conduct an ablation study over the testing data to determine the impact of the number of inducing points and training epochs, measured with the \gls{rmse} of each experiment.
\autoref{tab:model_comparison} shows that the RMSE and standard deviation grow with more inducing points.
This is explained by overfitting, as trained flights perform better but unseen flights perform significantly worse using more inducing points.
{\color{edits} Furthermore, because inference time scales quadratically with the number of inducing points, we selected Model 3, with 300 inducing points, for implementation.
In addition to achieving the best performance, it also offers the lowest inference time.
}

\begin{table}
    \centering
    \caption{Model Performance Comparison with and without Variance Scaler. Model 1 uses 1000 epochs, all others 750 epochs.}
    \label{tab:model_comparison}
    \begin{tabular}{@{}C{1.2cm}*{5}{c}@{}}
        \toprule
        \multirow{2}{*}{Model} & \multirow{2}{*}{\shortstack{Inducing\\Points}} & \multicolumn{2}{c}{No Variance Scaler} & \multicolumn{2}{c}{Variance Scaler} \\
        \cmidrule(lr){3-4} \cmidrule(lr){5-6}
        & & Avg RMSE & Std Dev & Avg RMSE & Std Dev \\
        \midrule
        {\color{edits} No Model (catenary)} &  -    &  0.1337 & 0.1211 & - & - \\\midrule
        Model 1 & 400  & 0.0836 & 0.1370 & 0.0783 & 0.1276 \\
        Model 2 & 400 &  0.0821 & 0.1348 & 0.0776 & 0.1270 \\
        Model 3 & 300 &  \textbf{0.0809} & \textbf{0.1331} & \textbf{0.0776} & \textbf{0.1255} \\
        Model 4 & 500 & 0.0829 & 0.1365 & 0.0780 & 0.1283 \\
        \bottomrule
    \end{tabular}
    \vspace{-0.5cm}
\end{table}

We choose a \gls{gp} over other machine learning models to have insight into whether the current state lies within the training regime during operation, using the variance of the Gaussian output.
If this is the case, we opt to scale down or completely disable the error compensation to avoid catastrophic estimation errors. 
In particular, we choose to scale the error compensation linearly with the output variance of the \gls{gp}, while ensuring hard maximums and minimums.
I.e., we {\color{edits} combine output of the \gls{gp} and the tether-localized position in a convex manner:}
\begin{align}
    \begin{aligned}
        \mathbf{x}_{d,comp} &= \mathbf{x}_d + \lambda\,\mathbf{e}_d,\\ 
        \text{where},\ 
        \lambda &= 1 - \frac{\text{\texttt{clamp}}(\sigma_{\text{GP}}, \sigma_{\min}, \sigma_{\max}) - \sigma_{\min}}{\sigma_{\max} - \sigma_{\min}}
    \end{aligned}
\end{align}
where $\sigma_{\text{GP}}$ is the variance of the \gls{gp} output.
We determine empirically the lower and upper thresholds at $\sigma_{\text{min}} = \SI{0.015}{\meter}$ and $\sigma_{\text{max}} = \SI{0.065}{\meter}$.
This provides safety guarantees and enables us to test the \gls{gp} in the closed-loop localization pipeline. {\color{edits} These thresholds correspond to the 10th and 90th percentiles of the localization error observed in preliminary runs of the standalone tether-based estimator, defining best- and worst-case performance baselines. The GP correction is fully applied when its predictive uncertainty falls below $\sigma_{\text{min}}$, indicating higher confidence than the standalone estimator can offer, and discarded when it exceeds $\sigma_{\text{max}}$, indicating lower reliability than the worst observed baseline.
Note that this provides a principled way to integrate the full measurement pipeline into conventional estimation frameworks.}

\vspace{-0.1cm}\section{System Design}\label{chap:hw}

This section details our design and fabrication of the \gls{tms} and the accompanying sensor suite, both developed by us specifically to support the localization pipeline. A crucial part, as the precise fabrication allows for low sensor uncertainties that directly contribute to position uncertainties. The integrated system fulfills two primary functions: 
\begin{enumerate}[label=\roman*.]
    \item Collecting tether state variables ($\alpha_o$, $\beta_o$, $\beta_d$, and $L$) required for localization
    \item Managing tether length and tension to prevent entanglement while preserving drone mobility within the maximum tether length. 
\end{enumerate}
To isolate the localization problem, the current implementation does not include data or power transfer between the platform and the drone.

\begin{table}
    \centering
    \caption{Arbitrary flight data point with sensor uncertainty computed from the standard deviation over \SI{10}{\second}.}
    \begin{tabular}{@{}l*{4}{C{1.45cm}}@{}}
    \toprule
    & $\alpha_o$ (rad) & $\beta_o$ (rad) & $\beta_d$ (rad) & $L$ (m) \\
    \midrule
    Sample & 1.774 & 0.841 & 1.117 & 4.284 \\
    Std. Dev. $\sigma$ & $\pm$0.0005& $\pm$0.0096 & $\pm$0.026 & $\pm$0.003 \\
    \bottomrule
    \end{tabular}
    \vspace{-0.5cm}
    \label{tab:uncertainty}
\end{table}

\subsection{Tether Sensor Suite}
The angle sensors measure the elevation and azimuth of the tether relative to both the drone and platform vertical axes. As shown in \autoref{fig:tms_prl_exp} (positions 1 and 2), both sensors employ a two-axis gimbal design with low-friction potentiometers on each axis, designed by us and partially inspired by~\cite{ugv_uav, lima_thesis}. The platform-mounted sensor allows the tether to traverse both rotation axes before entering the \gls{tms}. To reduce friction at low angles -- i.e., when the tether lies almost horizontal -- this sensor incorporates small rollers (\SI{7.5}{\milli\meter} diameter) mounted on metal pins. Both sensors are calibrated by taking orthographic pictures in-line with the axis while moving the tether in the axis plane. Using these pictures, we can determine which angle corresponds to which sensor value, allowing us to determine the linear fit between sensor values and tether angles. Using the drone's attitude via its \gls{imu}, the system compensates the sensed tether angle, such that the output tether angle is always w.r.t. the horizon.

The length sensor (position 4 in \autoref{fig:tms_prl_exp}) uses a dual-roller system where the upper roller connects to a rotary encoder, and the lower roller operates on a spring-actuated shaft. Two lateral springs create a pinching force against the upper roller, preventing slippage and ensuring accurate length measurement. This design also functions as a guide pulley, directing the tether from the variable tension sensor position to the fixed drum location.
Calibration involves measurement of dispensed tether length using a laser rangefinder while recording corresponding encoder values.

\begin{figure}
    \centering
    \def\svgwidth{\columnwidth}
    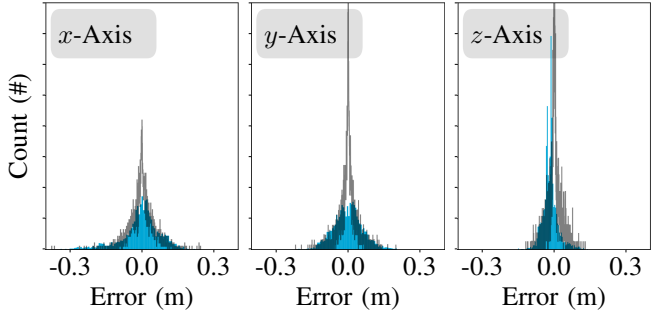
    \vspace{-0.75cm}
    \caption{Empirically observed (blue) and predicted (black) error distributions across 3000 samples for all estimation axes. The distributions clearly overlap, with only minor offsets in the $z$-axis, demonstrating the accuracy of our uncertainty prediction.}
    \label{fig:ideal_vs_observed}
    \vspace{-0.5cm}
\end{figure}

We evaluate the sensor standard deviation in static conditions in order to use these uncertainties with the partial derivatives from~\autoref{chap:uncertainty}. We report the results in~\autoref{tab:uncertainty}.
We then take 3000 random data points, compute the estimation error, and sample from the ideal state-dependent distribution using its covariance matrix $\boldsymbol{\Sigma}_\mathrm{output}$: 
\begin{align}
    \boldsymbol{\Sigma}_{\mathrm{output}} = \mathbf{J}\boldsymbol{\Sigma}_{\mathrm{input}}\mathbf{J}^\top
\end{align}
with $\mathbf{J}$ being the Jacobian of the estimation output with respect to the sensor input (i.e., the collection of the partial derivatives in Equations~\eqref{eq:uncertainty} to~\eqref{eq:uncertaintyend}) and $\boldsymbol{\Sigma}_\mathrm{input}$ denoting the input covariance matrix. 
As shown in~\autoref{fig:ideal_vs_observed}, these distributions mostly overlap, with only the $z$-Axis showing a small bias to underestimate the height.
This enables us to provide confidence intervals for the position estimation, which lets us integrate the estimate seamlessly into conventional estimator frameworks.
\subsection{Tether Management}
\begin{figure}
    \centering
    \begin{tikzpicture}

        \node[anchor=south west, inner sep=0] (base) at (0,0)
        {\ifdefined\anon
            \includegraphics[width=\columnwidth, trim={0 0 0 3.75cm}, clip]{media/highlighted_prl_anonymous.png}
            \else
            \includegraphics[width=\columnwidth, trim={0 1.5cm 0 3.75cm}, clip]{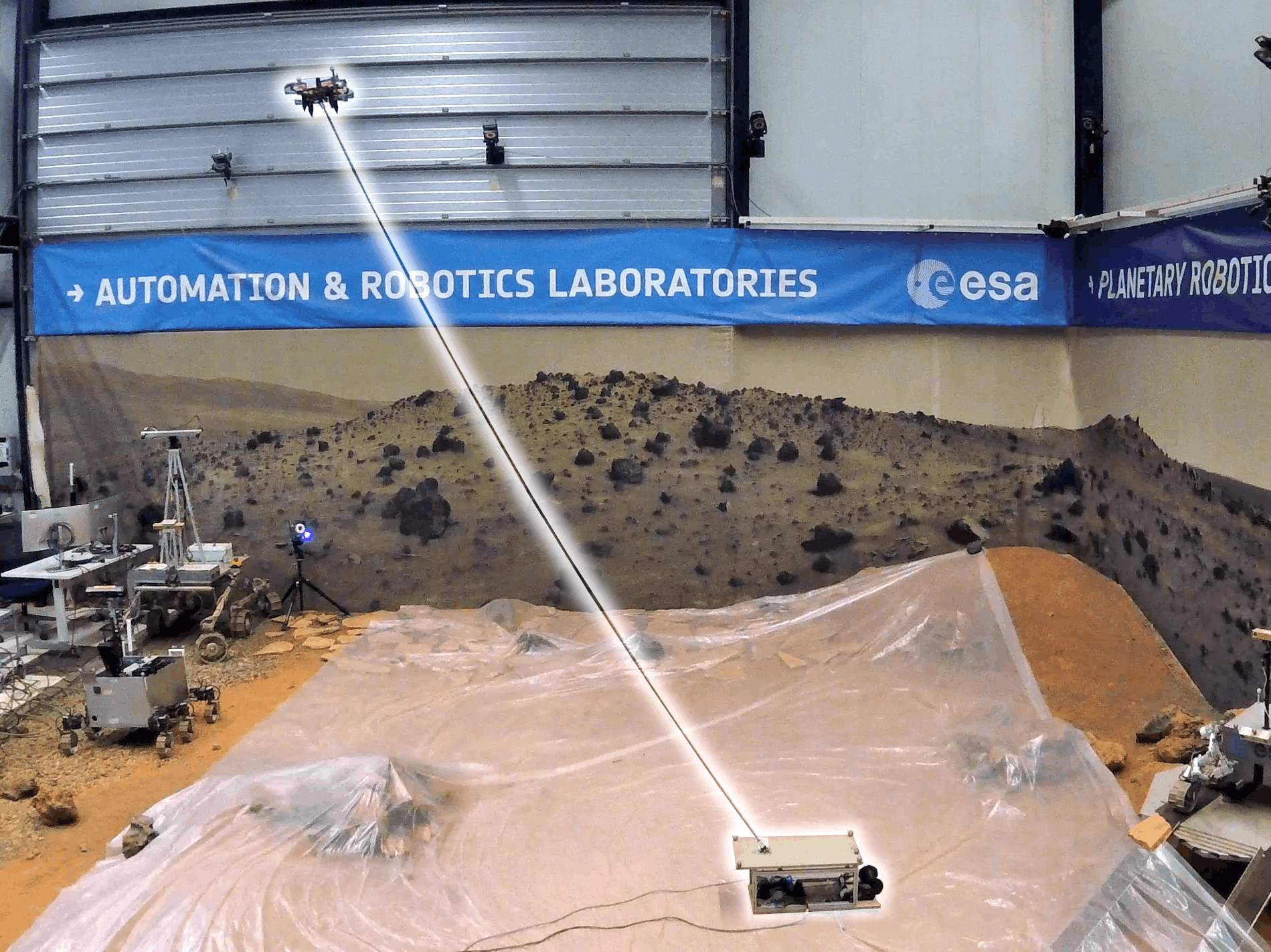}
            \fi
        };
      
        \def\baseW{5}   
        \def\baseH{5} 
      
        \coordinate (hl-sw) at (4.9, 0.05);   
        \coordinate (hl-ne) at (6.1, 0.75);   
        \coordinate (hl-se) at (6.1, 0.05);   
        
        \draw[delftblue, very thick, dashed, rounded corners=2pt]
          (hl-sw) rectangle (hl-ne);
      
        \def\insetW{4.75}   
        \def\insetH{3.78}   
      
        \coordinate (ins-sw) at (\baseW - \insetW + 3.475, \baseH - \insetH + 0.975);
      
        \node[anchor=south west, inner sep=0] (inset) at (ins-sw)
        {
            \renewcommand{\linewidth}{\insetW cm}
            \input{tikz/tms_prl.tex}
        };
      
        \coordinate (frame-sw) at ($ (inset.south west) + (0.11, 0.0) $);
        \coordinate (frame-ne) at ($ (inset.north east) - (0.11, 0.0) $);
        
        \draw[delftblue, ultra thick, dashed,rounded corners=2pt]

          (frame-sw) rectangle (frame-ne);
      
        \draw[delftblue, thick, dashed]
          (hl-se) -- ($ (frame-ne) - (0, \insetH)$);
      
    \end{tikzpicture}
    \vspace{-0.6cm}
    \caption{Localization experiment in the \prlname at the \estecname, with the \gls{tms} and the \gls{tuav} highlighted.
    Component 1. \& 2. highlight the drone and platform angle sensor, 3. is the tension sensor, 4. is the length sensor, and 5. is the tether drum. The tether is shown in yellow.}
    \vspace{-0.5cm}
    \label{fig:tms_prl_exp}
\end{figure}

{\color{edits}To support the localization pipeline, we include the \gls{tms} that regulates tether length to maintain constant tension across a wide range of \gls{uav} positions.
This ensures the catenary assumption holds, preventing buckling, looping, or ground contact, while also avoiding an overly taut elastic link.
This also keeps the angle sensors aligned with the true cable direction.
At the same time this additional tension needs to be compensated for by the \gls{uav} to maintain its position; in our case this tension is simply treated as an external disturbance, to be rejected by the position controller.
For these purposes} the tether drum assembly (position 5 in \autoref{fig:tms_prl_exp}) incorporates a stepper motor-driven system with a self-reversing leadscrew mechanism. This leadscrew provides uniform tether distribution across the drum width, preventing bunching and ensuring consistent operation. The tension sensor is designed with a spring-actuated arm (position 3 in \autoref{fig:tms_prl_exp}) that pivots around a lower axis with the upper end connected to a spring, as introduced by~\cite{lima_multi}. As tether tension increases, the arm rotates downward, with angular displacement measured via a potentiometer. This sensor is calibrated utilizing a spring balance across the \SIrange{0}{6}{\newton} range, establishing a quadratic relationship between potentiometer readings and tension.

\begin{figure}
    \centering
    \includegraphics[width=\columnwidth, trim={0 0.125cm 0 0}, clip]{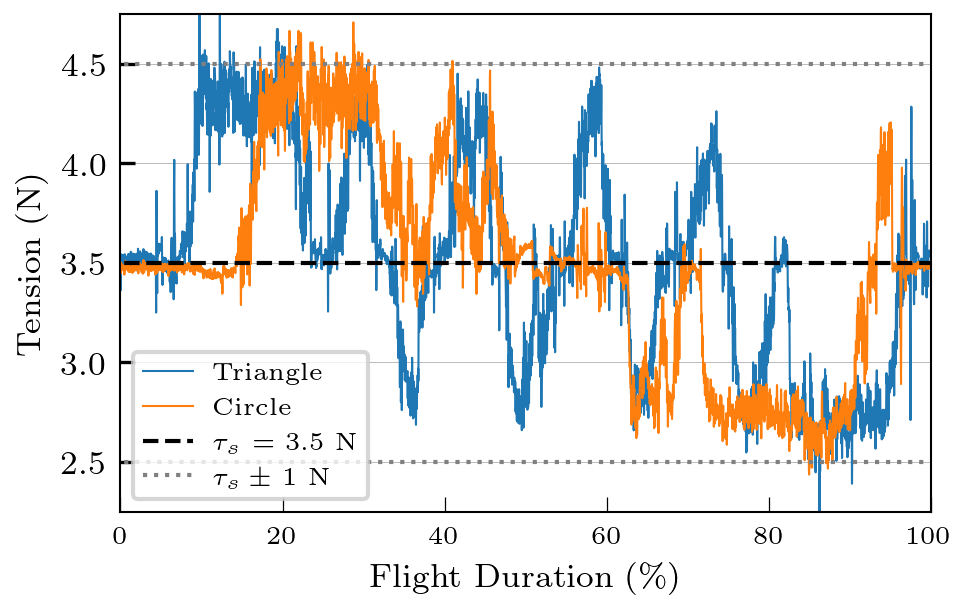}
    \vspace{-0.5cm}
    \caption{The tether tension for a triangular and circular flight experiment and its reference value $\tau_s$, regulated by the tension controller (\autoref{eq:tension-controller}) throughout a flight experiment. The controller successfully tracks the setpoint without causing large oscillations, staying within the reactivity threshold $\pm$\SI{1}{\newton}. The inflection points of the triangular flights can be seen causing extremes in tension.}
    \vspace{-0.5cm}
    \label{fig:tension-controller}
\end{figure}

A \gls{pd} controller maintains tether tension at \SI{3.5}{N}, determined empirically to balance drone mobility with system sensitivity.
The controller uses a low-pass filter for noise attenuation and implements a quadratic proportional term: reduced sensitivity for small deviations ($<$\SI{1}{\newton}) and enhanced responsiveness above:
\begin{align}
    \omega= k_P(\tau_s-\tau)|\tau_s-\tau|-k_D\dot{\tau}\label{eq:tension-controller}
\end{align}
where $\omega$ is the motor speed, $k_P$ and $k_D$ are the proportional and derivative gains, $\tau_s$ and $\tau$ are the tension setpoint and value. This prevents oscillations while ensuring rapid response to significant tension changes as shown in \autoref{fig:tension-controller}.

\subsection{System Integration}
The tether is a highly flexible silicone-insulated cable (\SI{22}{\kilo\gram\per\kilo\meter} density) to approximate the catenary model assumptions while providing sufficient mass to overcome sensor friction. We calculate the minimum required tether density based on the measured static angle sensor friction of \SI{0.029}{\newton} per sensor axis, yielding a required lower bound of \SI{15}{\gram\per\meter}, satisfied by the chosen tether. 
The mechanical framework utilizes 20mm aluminum extrusions with wooden platforms supporting all sensor components. Control electronics include an Arduino microcontroller for motor control and sensor interfacing, and a Raspberry Pi 5 with dedicated \gls{adc} for angle sensor acquisition and \gls{ros} communication.
The drone carries a Raspberry Pi 3A as a companion computer, interfacing with both the PX4 Autopilot and an onboard \gls{adc} for angle sensor data collection.
{\color{edits} The flight controller consumes the tether-based estimate and fuses it with \gls{imu} via an \gls{ekf}, to obtain high frequency state estimates. }

\vspace{-0.1cm}\section{Experiments}\label{chap:res}
To evaluate the performance of the proposed drone--rover {\color{edits}tether-inertial} localization approach, comprehensive flight experiments were conducted using a custom \gls{tms} and a HolyBro QAV250 quadrotor in the \prlname at the \estecname, as shown in \autoref{fig:tms_prl_exp}.

\subsection{Experimental Setup}
A VICON motion capture system provides ground truth positioning data, constraining the operational height to approximately \SI{4.5}{\meter} due to camera coverage limitations. However, please note that the motion capture system is only used for ground truth comparison and the drone's position feedback is solely supplied by our {\color{edits}tether-inertial} Localization framework. All results presented are the raw results before merging with the inertial odometry in the \gls{ekf} of the drone.
The experimental campaign comprises 17 flight missions totaling 37 minutes of flight data across {\color{edits}three} distinct trajectory geometries: circular, triangular, and figure-eight patterns.
We evaluate three altitude configurations: flights at \SI{2.0}{\meter} height with \SI{1.5}{\meter} characteristic dimensions (radius for circles, side length for triangles), flights at \SI{3.5}{\meter} height with \SI{2.5}{\meter} characteristic dimensions, and flights at \SI{2}{\meter} height with \SI{1.5}{\meter} characteristic dimensions with a sinusoidal altitude profile.
The tether's maximum length varies between \SI{3}{\meter} and \SI{4.5}{\meter} for the different sizes, respectively.
This systematic variation allows the assessment of both altitude and scale effects on localization accuracy. 
{\color{edits}Each flight mission followed a standardized protocol: 
Initiate a vertical takeoff to the target altitude, autonomously follow the prescribed trajectory at a constant speed of \SI{0.2}{\meter\per\second} and lastly, descent back onto the platform.
The approach is evaluated across 17 closed-loop flights: 11 using the {\color{edits}tether-inertial} localization and 6 using the \gls{gp}-enhanced variant.}
\subsection{Results}

\begin{figure}
    \centering
    \includegraphics[width=\linewidth, trim={0 0.125cm 0 0}, clip]{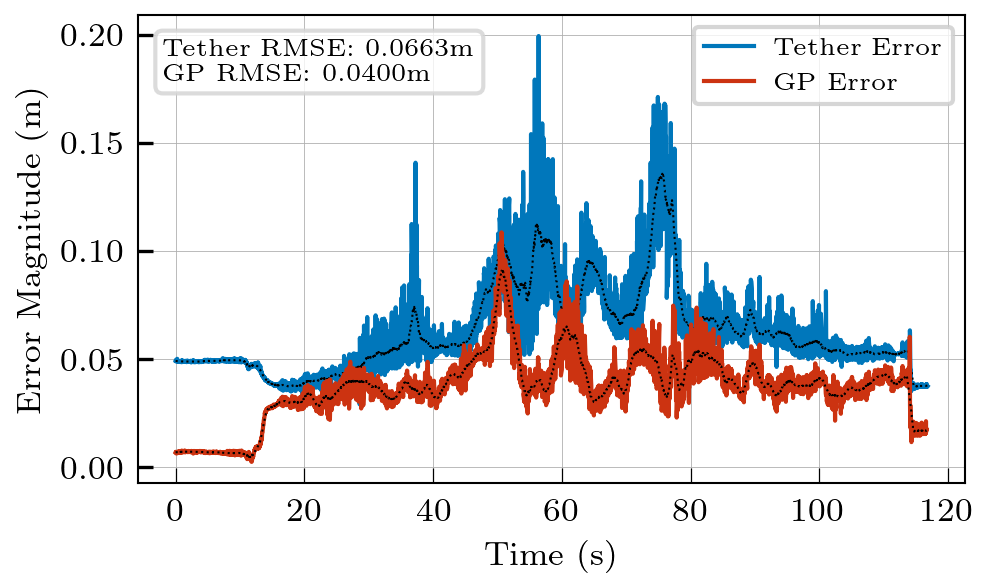}
    \vspace{-0.75cm}
    \caption{The magnitude of positional error using tether localization and \gls{gp} compensated error during a circle flight experiment of 2m high.}
    \label{fig:error_mag}
    \vspace{-0.75cm}
\end{figure}

\autoref{tab:localization_comparison} shows the estimation RMSE over all flight trajectories.
Hereby, we point out the difference between raw {\color{edits}tether-inertial} estimation and \gls{gp}-corrected estimation.
Across all experiments, we show that the raw {\color{edits}tether-inertial} position estimation achieves accuracies below \SI{10}{\centi\meter}.
Performance is better for trajectories with smaller characteristic dimensions, as shown in Equations \eqref{eq:uncertainty} to \eqref{eq:uncertaintyend}: the uncertainty scales linearly with the tether length $L$, which increases with characteristic dimension.
Furthermore, we demonstrate that the \gls{gp} method improves performance across all trials, achieving RMSEs up to \SI{2.2}{\centi\meter}. 
The overall improvement from {\color{edits}tether-inertial} localization (\SI{0.0744}{\meter} RMSE) to the \gls{gp}-enhanced {\color{edits}tether-inertial} system (\SI{0.0518}{\meter} average RMSE), represents a $30.3\%$ decrease of \gls{rmse} over the analytical model, and overall performance an order of magnitude better than previous works.
\begin{table}
    \centering
    \caption{Localization Performance Comparison Across Trajectories and Estimation Modes ({\color{edits}tether-inertial (TI)} and \gls{gp}). Average, worst, and best performances are bold in the last column.
    RMSEs {\color{edits}for trials with comparable characteristics} from related works are included for comparison.}
    \label{tab:localization_comparison}
    \vspace{-0.25cm}
    \renewcommand{\arraystretch}{1.1}
    \begin{tabular}
        {C{1.4cm}*{3}{C{0.9cm}}*{2}{C{1.075cm}}}
        \toprule
        Shape & Size (m) & Source & $n_\text{flights}$ & Duration (min) & RMSE (m) \\
        \midrule
        All & All & {\color{edits}TI} & $11$ & $25.0$ & $\mathbf{0.0744}$ \\
        All & All & \gls{gp} & $6$ & $11.8$ & $\mathbf{0.0518}$ \\
        \midrule
        \rowcolor{light-gray} Circle   & $1.5$ & TI & $4$ & $8.45$ & $0.0598$ \\
        \rowcolor{light-gray}          & $2.5$ & TI & $3$ &  $7.5$ & $\mathbf{0.0927}$ \\
        \rowcolor{light-gray}          & $1.5$ & \gls{gp} & $1$ &  $1.8$ & $0.0416$ \\
        \rowcolor{light-gray}          & $2.5$ & \gls{gp} & $1$ &  $2.0$ & $0.0649$ \\
                              Triangle & $1.5$ & TI & $1$ &  $2.3$ & $0.0544$ \\
                                       & $2.5$ & TI & $2$ &  $4.5$ & $0.0787$ \\
                                       & $1.5$ & \gls{gp} & $1$ &  $2.3$ & $\mathbf{0.0217}$ \\
                                       & $2.5$ & \gls{gp} & $1$ &  $2.6$ & $0.0572$ \\
        \rowcolor{light-gray} Eight    & $1.5$ & \gls{gp} & $2$ &  $3.2$ & $0.0566$ \\
        Circle (sinus. alt.) & $1.5$ & TI & $1$ &  $2.3$ & $0.0738$ \\
        \midrule\midrule
        \multicolumn{5}{l}{Xiao et al.~\cite{indoor_uav}} &  $0.37$ \\
        \multicolumn{5}{l}{Borgese et al.~\cite{ugv_uav}} &  $0.66$ \\
        \multicolumn{5}{l}{Lima et al.~\cite{lima_multi}} &  $1.1$ \\
        \multicolumn{5}{l}{Rico et al.~\cite{rico2021trajectory, rico2023analytics}} &  $0.95$ \\
        \bottomrule
    \end{tabular}
    \vspace{-0.25cm}
\end{table}
\autoref{fig:error_mag} illustrates this for a single flight. The \gls{gp} reduces the estimation error throughout the entire trajectory. Notably, at $t \approx \SI{58}{\second}$ and $t \approx \SI{78}{\second}$, the \gls{gp} compensates for large spikes in the raw {\color{edits}tether-inertial} estimates, highlighting its ability to correct for unmodeled effects.
{\color{edits} The second spike marks one of the trajectory's inflection points, where the \gls{uav} reverses toward the origin and induces tether behavior unmodelled by analytical model, but well compensated by the \gls{gp}.}
\begin{figure}
    \centering
    \includegraphics[width=\linewidth, trim={0 0.125cm 0 0.11cm}, clip]{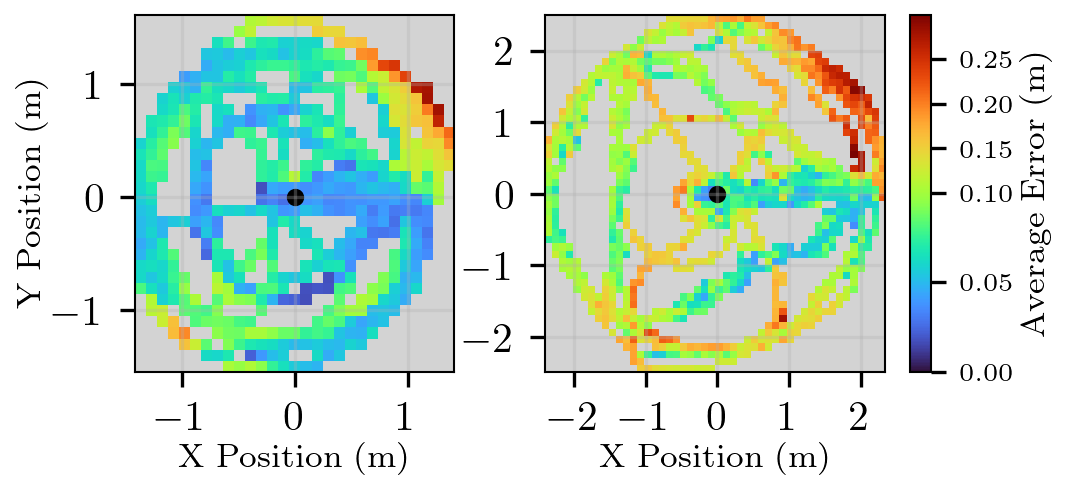}
    \vspace{-0.5cm}
    \caption{A top-down view of the average error for every 0.10m square around the origin (black). Left is the average error for flights of 1.5m characteristic dimension and right is for flights with a 2m dimension. Grey indicates that there are no samples in the voxel.}
    \label{fig:error_voxel}
    \vspace{-0.5cm}
\end{figure}
\begin{figure*}
    \centering
    \def\svgwidth{\linewidth}
    \vspace{-0.2cm}
    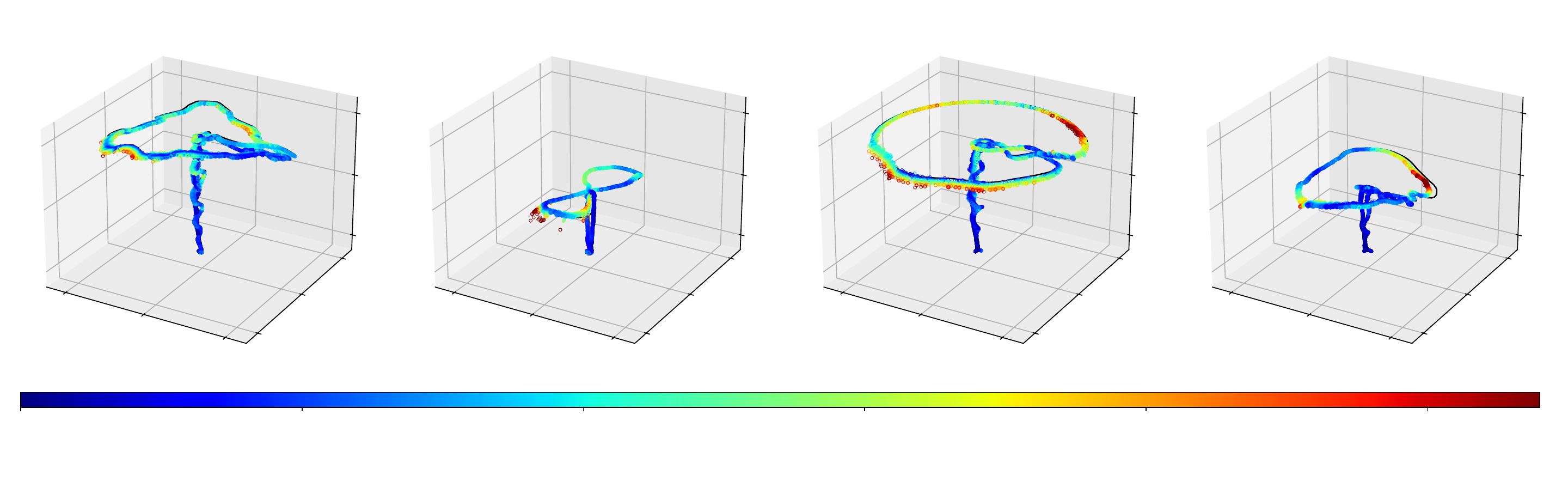
    \vspace{-0.75cm}
    \caption{Examples of different flight shapes used in the experiments. The tether-localized position is plotted in color according to the position error w.r.t. the ground truth flight path, which is plotted in black.
    We conduct experiments with triangular, circular, figure-eight, and sinusoidal height profile trajectory.}
    \label{fig:flight_error}
    \vspace{-0.25cm}
\end{figure*}

Similarly, \autoref{tab:localization_comparison} shows the influence of trajectory geometry and scale on localization accuracy. In particular, it shows that larger trajectory dimensions consistently increase localization error across all geometries. However, the system is able to provide localization estimates that are sufficient for position control ($\text{RMSE} \leq \SI{0.10}{\meter}$) in all cases.
The same trend can be observed in \autoref{fig:error_voxel}, which displays the mean error for \SI{10}{\centi\meter} voxels in the horizontal plane. 
We observe error-prone regions for the small and larger test flights that are concentrated particularly for large positive $x$ and $y$ positions.
It becomes especially clear in the right plot that the angle sensors deviate in the positive XY direction, sometimes causing errors of up to \SI{0.30}{\meter}.
This indicates a biased systematic error in one side of the angle sensor.
Lastly, we want to showcase tether configurations that introduce technical difficulties for state estimation. 
\autoref{fig:flight_error} illustrates example trials for the four selected trajectories, respectively, with position error magnitude color-coded relative to the ground truth position (shown in black). Notable deviations from the ground truth occur primarily at trajectory inflection points, particularly evident in the triangular pattern following sharp corners.
This behavior results from {\color{edits} two factors: excitation of the tether dynamics and the low bandwidth of the tether tension controller. 
At these inflection points the inertia of the tether causes it to diverge from the quasi-static assumption of the catenary model. At the same time the tension controller} requires a minimum amount of time to re-converge to the appropriate tension. In the transient phase, which correlates with the trajectory inflection points, overcoming the angle sensor friction becomes increasingly difficult as tether tension drops.
This violates the catenary assumptions as effectively, the end of the angle sensor acts as a new anchor point until the friction can be overcome.
The figure-eight trajectory exhibits the most challenging dynamics for the tether management system, requiring two rapid reductions in tether length per cycle where we can observe this behavior.
Hereby, tether tension reduces below the \SI{3.5}{\newton} setpoint, causing momentary degradation in angle sensor accuracy as the tether approaches slack conditions.
Finally, compared to related works (\autoref{tab:localization_comparison}), our system achieves up to ten times better RMSE accuracy, outperforming the state-of-the-art by at least a factor of four in all cases.
\vspace{-0.15cm}\section{Conclusion}
This work presents a drone-rover tether-inertial localization pipeline for centimetre-level positioning accuracy of tethered drones. In GNSS-denied environments or in scenarios where other methods, such as visual-inertial odometry, struggle, this approach provides a reliable alternative. Our approach combines a computationally efficient analytical catenary model with Gaussian Process error compensation to achieve accurate real-time and non-drifting localization, thoroughly tested on a custom developed \gls{tms}, while also providing uncertainty guarantees from the sensor accuracies.
The analytical catenary model provides closed-form solutions for estimating the drone's position from tether state measurements. These measurements are obtained using our custom \gls{tms}, integrated with the customized drone. Experimental validation of the analytical model over 25 minutes of flight demonstrated an average RMSE of \SI{0.0744}{\meter}.
Test flights included autonomous circular, triangular, and figure-eight trajectories to evaluate performance across diverse motion patterns.
The integration of a \gls{gp}, trained on pairs of tether states and residual errors to the \gls{mocap} obtained ground truth positions, reduced localization error by $30.3\%$, achieving an average RMSE of \SI{0.0518}{\meter} over 12 additional minutes with variance-based scaling of residual error compensation. 
The demonstrated localization accuracy of approximately \SI{5}{\centi\meter} represents an order of magnitude improvement over existing \gls{tuav} localization methods and enables autonomous flight operations in feature-sparse environments, such as planetary surfaces, with high accuracy.
In future work we intend to extend the estimation methodology to also infer the heading of the \gls{tuav} using an orthogonal angle sensor at the anchor point, incorporating a tether dynamic model {\color{edits} and a \gls{gp} with higher order inputs to provide estimates of the \gls{uav} velocities and improve the accuracy in highly dynamic regimes.}.
Future work will focus on integrating this approach into a mobile rover with active {\color{edits} feed-forward tether control, enabling the drone and \gls{tms} to cooperatively regulate tether length to avoid dynamic tether transient behaviors and consequently improving the estimation accuracy.
At the same time, improves space-grade sensor will allow for operation at higher altitudes and larger tether lengths}. This will create a robust rover-drone system suitable for challenging planetary exploration tasks.
\vspace{-0.1cm}
\printbibliography
\end{document}

%% file: tikz/catenary.tex
\begin{tikzpicture}
  \begin{axis}[
    axis lines=center,
    xmin=0.0, xmax=3.3,
    ymin=0.0, ymax=3.3,
    zmin=0.0, zmax=2.5,
    clip=false, 
    xlabel={$r$},
    x label style={at={(axis description cs:0.5,-0.02)},anchor=north},
    y label style={at={(axis description cs:0.5, 0.3)},anchor=west},
    ylabel={$r_\bot$},
    zlabel={$z$},
    xticklabel=\empty,
    yticklabel=\empty,
    zticklabel=\empty,
    extra x ticks={3.29},
    extra y ticks={3.29},
    view={45}{15},    
  ]
    \addplot3[
        thick,
        domain=0:3.5,
        samples=30,
        samples y = 0,
        fill=none
    ] ({x},{x},{3.9392*cosh((x+0.6910)/3.93923) - 4});
    
    \node[circle, fill=red, inner sep=2.0pt] at (axis cs:0,0,0) (O) {};
    \node[inner sep=2.0pt] at (axis cs:3.29,3.29,2.13){
      \includegraphics[width=1.0cm]{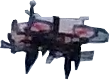}
    };
    \node[circle, fill=green, inner sep=2.0pt] at (axis cs:3.29,3.29,2.13) (P) {};
    
    \node[red, below left=3pt] at (O) {$o$};
    \node[green, below right=2pt] at (P) {$d$};

    \draw[gray, loosely dotted] (0, 3.29, 0) -- (3.29, 3.29, 0);
    \draw[gray, loosely dotted] (3.29, 0, 0) -- (3.29, 3.29, 0);
    \draw[gray, dashed] (0, 0, 0) -- (3.29, 3.29, 0);
    \draw[gray, loosely dotted] (P) -- (3.29, 3.29, 0) node[midway, right, black] {$z_d$};

    \node[gray] (H0) at (2.0, 0.0, 0) {};
    \node[gray] (H1) at (2.0, 2.0, 0) {};
    \draw[gray, dashed] (O) -- ++(2.0, 2.0, 0.4) coordinate (T1);
    
    \draw[gray, dashed] (P) -- ++(0.5, 0.5, 0.6) coordinate (T2);
    \draw[gray, dashed] (P) -- ++(0.5, 0.5, 0) coordinate (H2);
    
    \pic [draw, ->, black, angle eccentricity=1.1, angle radius=4cm, "$\beta_o$"] {angle = {H1--O--T1}};   
    \pic [draw, ->, black, angle eccentricity=1.4, angle radius=0.7cm, "$\beta_d$"] {angle = {H2--P--T2}};
    \pic [draw, ->, black, angle eccentricity=1.1, angle radius=3.0cm, "$\alpha_o$"] {angle = {H0--O--H1}};
    
    \addplot3[samples=80,
            samples y=0, 
            thick, blue, domain=0.0:3.2,
            decoration={markings,
                    mark=at position 0.0 with {\arrowreversed{stealth}},
                    mark=at position 1.0 with {\arrow{stealth}}},
            postaction={decorate}
    ]
      ({x}, {x}, {3.9392*cosh((x+0.6910)/3.83923) - 4 + 0.25});
    
    \node[blue, above] at (2.9, 2.9, 2.1) {$L$};
    
  \end{axis}
\end{tikzpicture}

%% file: media/ideal_vs_observed_error_dist.eps_tex
\begingroup%
  \makeatletter%
  \providecommand\color[2][]{%
    \errmessage{(Inkscape) Color is used for the text in Inkscape, but the package 'color.sty' is not loaded}%
    \renewcommand\color[2][]{}%
  }%
  \providecommand\transparent[1]{%
    \errmessage{(Inkscape) Transparency is used (non-zero) for the text in Inkscape, but the package 'transparent.sty' is not loaded}%
    \renewcommand\transparent[1]{}%
  }%
  \providecommand\rotatebox[2]{#2}%
  \newcommand*\fsize{\dimexpr\f@size pt\relax}%
  \newcommand*\lineheight[1]{\fontsize{\fsize}{#1\fsize}\selectfont}%
  \ifx\svgwidth\undefined%
    \setlength{\unitlength}{648bp}%
    \ifx\svgscale\undefined%
      \relax%
    \else%
      \setlength{\unitlength}{\unitlength * \real{\svgscale}}%
    \fi%
  \else%
    \setlength{\unitlength}{\svgwidth}%
  \fi%
  \global\let\svgwidth\undefined%
  \global\let\svgscale\undefined%
  \makeatother%
  \begin{picture}(1,0.5)%
    \lineheight{1}%
    \setlength\tabcolsep{0pt}%
    \put(0,0){\includegraphics[width=\unitlength]{ideal_vs_observed_error_dist.eps}}%
    \put(0.09515432,0.06691601){\makebox(0,0)[t]{\lineheight{1.25}\smash{\begin{tabular}[t]{c}-0.3\end{tabular}}}}%
    \put(0.20524691,0.06691601){\makebox(0,0)[t]{\lineheight{1.25}\smash{\begin{tabular}[t]{c}0.0\end{tabular}}}}%
    \put(0.3153395,0.06691601){\makebox(0,0)[t]{\lineheight{1.25}\smash{\begin{tabular}[t]{c}0.3\end{tabular}}}}%
    \put(0.20524691,0.02111642){\makebox(0,0)[t]{\lineheight{1.25}\smash{\begin{tabular}[t]{c}Error (m)\end{tabular}}}}%
    \put(0.02901283,0.29526235){\rotatebox{90}{\makebox(0,0)[t]{\lineheight{1.25}\smash{\begin{tabular}[t]{c}Count (\#)\end{tabular}}}}}%
    \put(0.07680556,0.42265374){\makebox(0,0)[lt]{\lineheight{1.25}\smash{\begin{tabular}[t]{l}$x$-Axis\end{tabular}}}}%
    \put(0.41080249,0.06691601){\makebox(0,0)[t]{\lineheight{1.25}\smash{\begin{tabular}[t]{c}-0.3\end{tabular}}}}%
    \put(0.52089507,0.06691601){\makebox(0,0)[t]{\lineheight{1.25}\smash{\begin{tabular}[t]{c}0.0\end{tabular}}}}%
    \put(0.63098766,0.06691601){\makebox(0,0)[t]{\lineheight{1.25}\smash{\begin{tabular}[t]{c}0.3\end{tabular}}}}%
    \put(0.52089507,0.02111642){\makebox(0,0)[t]{\lineheight{1.25}\smash{\begin{tabular}[t]{c}Error (m)\end{tabular}}}}%
    \put(0.3924537,0.42265374){\makebox(0,0)[lt]{\lineheight{1.25}\smash{\begin{tabular}[t]{l}$y$-Axis\end{tabular}}}}%
    \put(0.7264506,0.06691601){\makebox(0,0)[t]{\lineheight{1.25}\smash{\begin{tabular}[t]{c}-0.3\end{tabular}}}}%
    \put(0.83654324,0.06691601){\makebox(0,0)[t]{\lineheight{1.25}\smash{\begin{tabular}[t]{c}0.0\end{tabular}}}}%
    \put(0.94663578,0.06691601){\makebox(0,0)[t]{\lineheight{1.25}\smash{\begin{tabular}[t]{c}0.3\end{tabular}}}}%
    \put(0.83654324,0.02111642){\makebox(0,0)[t]{\lineheight{1.25}\smash{\begin{tabular}[t]{c}Error (m)\end{tabular}}}}%
    \put(0.70810186,0.42265374){\makebox(0,0)[lt]{\lineheight{1.25}\smash{\begin{tabular}[t]{l}$z$-Axis\end{tabular}}}}%
  \end{picture}%
\endgroup%

%% file: tikz/tms_prl.tex
\begin{tikzpicture}
    \node[anchor=south west,inner sep=0] (image) at (0,0) {\includegraphics[width=\linewidth]{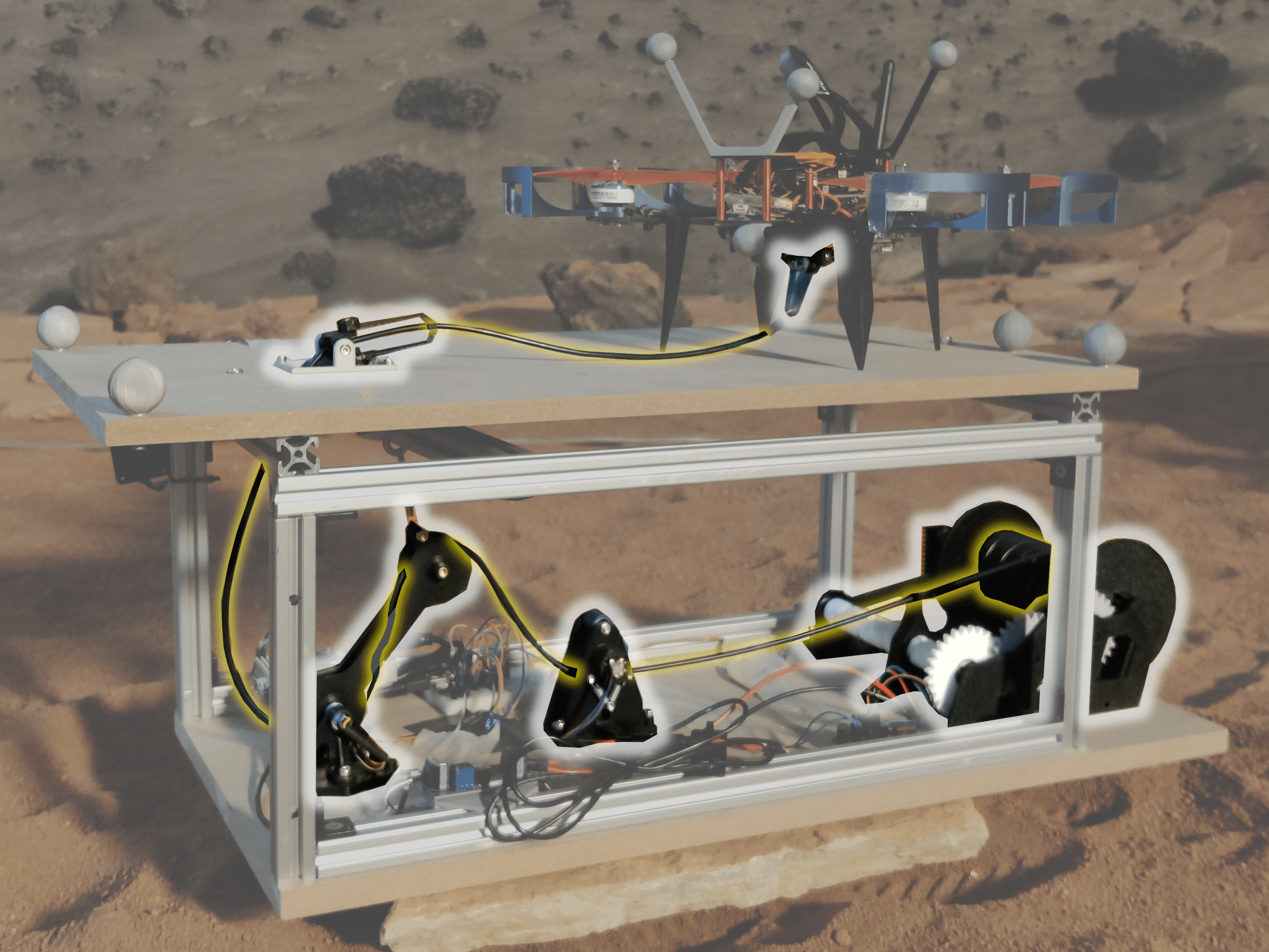}};
    
    \begin{scope}[x={(image.south east)},y={(image.north west)}]
        
        \node[fill=white, fill opacity=0.7, text=black, font=\bfseries\large, 
              rounded corners=3pt, minimum width=0.6cm, minimum height=0.6cm] 
              at (0.62, 0.75) {$1$};
        
        \node[fill=white, fill opacity=0.7, text=black, font=\bfseries\large, 
              rounded corners=3pt, minimum width=0.6cm, minimum height=0.6cm] 
              at (0.15, 0.75) {$2$};
        
        \node[fill=white, fill opacity=0.7, text=black, font=\bfseries\large, 
              rounded corners=3pt, minimum width=0.6cm, minimum height=0.6cm] 
              at (0.12, 0.35) {$3$};
        
        \node[fill=white, fill opacity=0.7, text=black, font=\bfseries\large, 
              rounded corners=3pt, minimum width=0.6cm, minimum height=0.6cm] 
              at (0.45, 0.20) {$4$};
        
        \node[fill=white, fill opacity=0.7, text=black, font=\bfseries\large, 
              rounded corners=3pt, minimum width=0.6cm, minimum height=0.6cm] 
              at (0.80, 0.20) {$5$};
    \end{scope}
\end{tikzpicture}

%% file: media/flight_path_error.eps_tex
\begingroup%
  \makeatletter%
  \providecommand\color[2][]{%
    \errmessage{(Inkscape) Color is used for the text in Inkscape, but the package 'color.sty' is not loaded}%
    \renewcommand\color[2][]{}%
  }%
  \providecommand\transparent[1]{%
    \errmessage{(Inkscape) Transparency is used (non-zero) for the text in Inkscape, but the package 'transparent.sty' is not loaded}%
    \renewcommand\transparent[1]{}%
  }%
  \providecommand\rotatebox[2]{#2}%
  \newcommand*\fsize{\dimexpr\f@size pt\relax}%
  \newcommand*\lineheight[1]{\fontsize{\fsize}{#1\fsize}\selectfont}%
  \ifx\svgwidth\undefined%
    \setlength{\unitlength}{1381.52996546bp}%
    \ifx\svgscale\undefined%
      \relax%
    \else%
      \setlength{\unitlength}{\unitlength * \real{\svgscale}}%
    \fi%
  \else%
    \setlength{\unitlength}{\svgwidth}%
  \fi%
  \global\let\svgwidth\undefined%
  \global\let\svgscale\undefined%
  \makeatother%
  \begin{picture}(1,0.31472932)%
    \lineheight{1}%
    \setlength\tabcolsep{0pt}%
    \put(0,0){\includegraphics[width=\unitlength]{flight_path_error.pdf}}%
    \put(0.0303824,0.10630967){\color[rgb]{0,0,0}\makebox(0,0)[lt]{\lineheight{1.25}\smash{\begin{tabular}[t]{l}-2\end{tabular}}}}%
    \put(0.08131275,0.09179678){\color[rgb]{0,0,0}\makebox(0,0)[lt]{\lineheight{1.25}\smash{\begin{tabular}[t]{l}0\end{tabular}}}}%
    \put(0.13347593,0.07662085){\color[rgb]{0,0,0}\makebox(0,0)[lt]{\lineheight{1.25}\smash{\begin{tabular}[t]{l}2\end{tabular}}}}%
    \put(0.17625531,0.08256209){\color[rgb]{0,0,0}\makebox(0,0)[lt]{\lineheight{1.25}\smash{\begin{tabular}[t]{l}-2\end{tabular}}}}%
    \put(0.20471361,0.10798245){\color[rgb]{0,0,0}\makebox(0,0)[lt]{\lineheight{1.25}\smash{\begin{tabular}[t]{l}0\end{tabular}}}}%
    \put(0.23013109,0.13155415){\color[rgb]{0,0,0}\makebox(0,0)[lt]{\lineheight{1.25}\smash{\begin{tabular}[t]{l}2\end{tabular}}}}%
    \put(0.24369141,0.16298524){\color[rgb]{0,0,0}\makebox(0,0)[lt]{\lineheight{1.25}\smash{\begin{tabular}[t]{l}0\end{tabular}}}}%
    \put(0.24717668,0.24042257){\color[rgb]{0,0,0}\makebox(0,0)[lt]{\lineheight{1.25}\smash{\begin{tabular}[t]{l}4\end{tabular}}}}%
    \put(0.01326993,0.28500000){\color[rgb]{0,0,0}\makebox(0,0)[lt]{\lineheight{1.25}\smash{\begin{tabular}[t]{l}Triangle Flight Path\end{tabular}}}}%
    \put(0.27835805,0.10630967){\color[rgb]{0,0,0}\makebox(0,0)[lt]{\lineheight{1.25}\smash{\begin{tabular}[t]{l}-2\end{tabular}}}}%
    \put(0.32928854,0.09179678){\color[rgb]{0,0,0}\makebox(0,0)[lt]{\lineheight{1.25}\smash{\begin{tabular}[t]{l}0\end{tabular}}}}%
    \put(0.38145172,0.07662085){\color[rgb]{0,0,0}\makebox(0,0)[lt]{\lineheight{1.25}\smash{\begin{tabular}[t]{l}2\end{tabular}}}}%
    \put(0.42423111,0.08256209){\color[rgb]{0,0,0}\makebox(0,0)[lt]{\lineheight{1.25}\smash{\begin{tabular}[t]{l}-2\end{tabular}}}}%
    \put(0.45268941,0.10798245){\color[rgb]{0,0,0}\makebox(0,0)[lt]{\lineheight{1.25}\smash{\begin{tabular}[t]{l}0\end{tabular}}}}%
    \put(0.47810688,0.13155415){\color[rgb]{0,0,0}\makebox(0,0)[lt]{\lineheight{1.25}\smash{\begin{tabular}[t]{l}2\end{tabular}}}}%
    \put(0.49166721,0.16298524){\color[rgb]{0,0,0}\makebox(0,0)[lt]{\lineheight{1.25}\smash{\begin{tabular}[t]{l}0\end{tabular}}}}%
    \put(0.49515248,0.24042257){\color[rgb]{0,0,0}\makebox(0,0)[lt]{\lineheight{1.25}\smash{\begin{tabular}[t]{l}4\end{tabular}}}}%
    \put(0.26124587,0.28500000){\color[rgb]{0,0,0}\makebox(0,0)[lt]{\lineheight{1.25}\smash{\begin{tabular}[t]{l}Eight Flight Path\end{tabular}}}}%
    \put(0.52633385,0.10630967){\color[rgb]{0,0,0}\makebox(0,0)[lt]{\lineheight{1.25}\smash{\begin{tabular}[t]{l}-2\end{tabular}}}}%
    \put(0.57726434,0.09179678){\color[rgb]{0,0,0}\makebox(0,0)[lt]{\lineheight{1.25}\smash{\begin{tabular}[t]{l}0\end{tabular}}}}%
    \put(0.62942752,0.07662085){\color[rgb]{0,0,0}\makebox(0,0)[lt]{\lineheight{1.25}\smash{\begin{tabular}[t]{l}2\end{tabular}}}}%
    \put(0.6722069,0.08256209){\color[rgb]{0,0,0}\makebox(0,0)[lt]{\lineheight{1.25}\smash{\begin{tabular}[t]{l}-2\end{tabular}}}}%
    \put(0.7006652,0.10798245){\color[rgb]{0,0,0}\makebox(0,0)[lt]{\lineheight{1.25}\smash{\begin{tabular}[t]{l}0\end{tabular}}}}%
    \put(0.72607906,0.13155415){\color[rgb]{0,0,0}\makebox(0,0)[lt]{\lineheight{1.25}\smash{\begin{tabular}[t]{l}2\end{tabular}}}}%
    \put(0.73964373,0.16298524){\color[rgb]{0,0,0}\makebox(0,0)[lt]{\lineheight{1.25}\smash{\begin{tabular}[t]{l}0\end{tabular}}}}%
    \put(0.74312538,0.24042257){\color[rgb]{0,0,0}\makebox(0,0)[lt]{\lineheight{1.25}\smash{\begin{tabular}[t]{l}4\end{tabular}}}}%
    \put(0.50922166,0.28500000){\color[rgb]{0,0,0}\makebox(0,0)[lt]{\lineheight{1.25}\smash{\begin{tabular}[t]{l}Circle Flight Path\end{tabular}}}}%
    \put(0.77430819,0.10630967){\color[rgb]{0,0,0}\makebox(0,0)[lt]{\lineheight{1.25}\smash{\begin{tabular}[t]{l}-2\end{tabular}}}}%
    \put(0.82523724,0.09179678){\color[rgb]{0,0,0}\makebox(0,0)[lt]{\lineheight{1.25}\smash{\begin{tabular}[t]{l}0\end{tabular}}}}%
    \put(0.87740404,0.07662085){\color[rgb]{0,0,0}\makebox(0,0)[lt]{\lineheight{1.25}\smash{\begin{tabular}[t]{l}2\end{tabular}}}}%
    \put(0.9201827,0.08256209){\color[rgb]{0,0,0}\makebox(0,0)[lt]{\lineheight{1.25}\smash{\begin{tabular}[t]{l}-2\end{tabular}}}}%
    \put(0.94864389,0.10798245){\color[rgb]{0,0,0}\makebox(0,0)[lt]{\lineheight{1.25}\smash{\begin{tabular}[t]{l}0\end{tabular}}}}%
    \put(0.97405775,0.13155415){\color[rgb]{0,0,0}\makebox(0,0)[lt]{\lineheight{1.25}\smash{\begin{tabular}[t]{l}2\end{tabular}}}}%
    \put(0.98762242,0.16298524){\color[rgb]{0,0,0}\makebox(0,0)[lt]{\lineheight{1.25}\smash{\begin{tabular}[t]{l}0\end{tabular}}}}%
    \put(0.99110407,0.24042257){\color[rgb]{0,0,0}\makebox(0,0)[lt]{\lineheight{1.25}\smash{\begin{tabular}[t]{l}4\end{tabular}}}}%
    \put(0.75719673,0.28500000){\color[rgb]{0,0,0}\makebox(0,0)[lt]{\lineheight{1.25}\smash{\begin{tabular}[t]{l}Path with Altitude Changes\end{tabular}}}}%
    \put(0.00521161,0.03238627){\color[rgb]{0,0,0}\makebox(0,0)[lt]{\lineheight{1.25}\smash{\begin{tabular}[t]{l}0.00\end{tabular}}}}%
    \put(0.18472273,0.03238627){\color[rgb]{0,0,0}\makebox(0,0)[lt]{\lineheight{1.25}\smash{\begin{tabular}[t]{l}0.05\end{tabular}}}}%
    \put(0.36423386,0.03238627){\color[rgb]{0,0,0}\makebox(0,0)[lt]{\lineheight{1.25}\smash{\begin{tabular}[t]{l}0.10\end{tabular}}}}%
    \put(0.54374498,0.03238627){\color[rgb]{0,0,0}\makebox(0,0)[lt]{\lineheight{1.25}\smash{\begin{tabular}[t]{l}0.15\end{tabular}}}}%
    \put(0.7232561,0.03238627){\color[rgb]{0,0,0}\makebox(0,0)[lt]{\lineheight{1.25}\smash{\begin{tabular}[t]{l}0.20\end{tabular}}}}%
    \put(0.90276722,0.03238627){\color[rgb]{0,0,0}\makebox(0,0)[lt]{\lineheight{1.25}\smash{\begin{tabular}[t]{l}0.25\end{tabular}}}}%
    \put(0.48161314,0.00671584){\color[rgb]{0,0,0}\makebox(0,0)[lt]{\lineheight{1.25}\smash{\begin{tabular}[t]{l}Error (m)\end{tabular}}}}%
  \end{picture}%
\endgroup%

%% file: references.bib
@article{balaram2021ingenuity,
  title={The ingenuity helicopter on the perseverance rover},
  author={Balaram, J and Aung, MiMi and Golombek, Matthew P},
  journal={Space Science Reviews},
  volume={217},
  number={4},
  %pages={56},
  year={2021},
  publisher={Springer}
}

@article{mangold2021perseverance,
  title={Perseverance rover reveals an ancient delta-lake system and flood deposits at Jezero crater, Mars},
  author={Mangold, Nicolas and Gupta, S and Gasnault, O and Dromart, G and Tarnas, JD and Sholes, SF and Horgan, B and Quantin-Nataf, C and Brown, AJ and Le Mou{\'e}lic, St{\'e}phane and others},
  journal={Science},
  volume={374},
  number={6568},
  %%pages={711--717},
  year={2021},
  publisher={American Association for the Advancement of Science}
}

@misc{NASA_2024, author={DC Agle}, url={https://www.jpl.nasa.gov/news/nasa-performs-first-aircraft-accident-investigation-on-another-world/}, journal={NASA}, publisher={NASA}, year={2024}}

@article{Folorunsho2024-bj,
  title={Redefining aerial innovation: autonomous tethered drones as a solution to battery life and data latency challenges},
  author={Folorunsho, Samuel O and Norris, William R},
  journal={arXiv preprint arXiv:2403.07922},
  year={2024}
}

@BOOK{Lockwood2007-wn,
  title     = "Book of Curves",
  author    = "Lockwood, E H",
  publisher = "Cambridge University Press",
  month     =  dec,
  year      =  2007,
  address   = "Cambridge, England"
}

@inproceedings{rico2021trajectory,
  title={Trajectory selection for power-over-tether atmospheric sensing UAS},
  author={Rico, Daniel A and Mu{\~n}oz-Arriola, Francisco and Detweiler, Carrick},
  booktitle={International Conference on Intelligent Robots and Systems},
  %pages={2321--2328},
  year={2021},
  %organization={IEEE}
}

@ARTICLE{dantonio2021catenary,
  author={D'Antonio, Diego S. and Cardona, Gustavo A. and Saldaña, David},
  journal={IEEE Robotics and Automation Letters}, 
  title={The Catenary Robot: Design and Control of a Cable Propelled by Two Quadrotors}, 
  year={2021},
  volume={6},
  number={2},
  %pages={3857-3863},
  keywords={Robots;Robot kinematics;Shape;Transportation;Trajectory;Unmanned aerial vehicles;Service robots;Aerial systems;applications;cellular and modular robots;mobile manipulation},
  %doi={10.1109/LRA.2021.3062603}
}

@inproceedings{rico2023analytics,
  title={Analytics for real-time inertial localization of the tethered aircraft unmanned system},
  author={Rico, Daniel A and Mu{\~n}oz-Arriola, Francisco and Bradley, Justin M and Detweiler, Carrick J},
  booktitle={International Symposium on Experimental Robotics},
  %pages={469--480},
  year={2023},
  %organization={Springer}
}

@BOOK{Rasmussen2005-ep,
  title     = "Gaussian processes for machine learning",
  author    = "Rasmussen, Carl Edward and Williams, Christopher K I",
  publisher = "MIT Press",
  series    = "Adaptive Computation and Machine Learning Series",
  month     =  jun,
  year      =  2005,
  address   = "London, England",
  language  = "en"
}

@inproceedings{mcintire2016sparse,
  title={Sparse Gaussian processes for Bayesian optimization.},
  author={McIntire, Mitchell and Ratner, Daniel and Ermon, Stefano},
  booktitle={UAI},
  volume={16},
  %pages={517--526},
  year={2016}
}

@article{kmeans,
title = {K-means clustering algorithms: A comprehensive review, variants analysis, and advances in the era of big data},
journal = {Information Sciences},
volume = {622},
%pages = {178-210},
year = {2023},
author = {Abiodun M. Ikotun and Absalom E. Ezugwu and Laith Abualigah and Belal Abuhaija and Jia Heming},
}

@inproceedings{boukoberine2019power,
  title={Power supply architectures for drones-a review},
  author={Boukoberine, Mohamed Nadir and Zhou, Zhibin and Benbouzid, Mohamed},
  booktitle={45th Annual Conference of the IEEE Industrial Electronics Society},
  volume={1},
  %pages={5826--5831},
  year={2019},
  %%organization={IEEE}
}

@article{motlagh2017uav,
  title={UAV-based IoT platform: A crowd surveillance use case},
  author={Motlagh, Naser Hossein and Bagaa, Miloud and Taleb, Tarik},
  journal={Communications Magazine},
  %volume={55},
  %number={2},
  %%pages={128--134},
  year={2017},
  publisher={IEEE}
}

@INPROCEEDINGS{10109426,
  author={Omandam, Reza S. and Paradela, Immanuel P. and Banglos, Charles Alver G. and Librado, Lester G. and Mae Canlas, Rocyle and Salaan, Carl John O.},
  booktitle={International Conference on Humanoid, Nanotechnology, Information Technology, Communication and Control, Environment, and Management}, 
  title={3D Localization of Suspended and Tethered Drone for High-rise Bridge Inspection}, 
  year={2022},
  volume={},
  number={},
  %pages={1-6},
  %keywords={Bridges;Location awareness;Training;Three-dimensional displays;Wires;Neural networks;Force;suspended and tethered-drone;localization;bridge inspection},
}

@Article{oxpecker,
AUTHOR = {Martinez Rocamora, Bernardo and Lima, Rogério R. and Samarakoon, Kieren and Rathjen, Jeremy and Gross, Jason N. and Pereira, Guilherme A. S.},
TITLE = {Oxpecker: A Tethered UAV for Inspection of Stone-Mine Pillars},
JOURNAL = {Drones},
VOLUME = {7},
YEAR = {2023},
NUMBER = {2},
ARTICLE-NUMBER = {73},
}

@article{lima_vectors,
author = {Lima, Rogerio and Martinez, Bernardo and Pereira, Guilherme},
year = {2023},
pages = {1-8},
title = {Continuous Vector Fields for Precise Cable-Guided Landing of Tethered UAVs},
journal = {IEEE Robotics and Automation Letters},
}

@Article{drones_review,
AUTHOR = {Fattori, Francesco and Cocuzza, Silvio},
TITLE = {Tethered Drones: A Comprehensive Review of Technologies, Challenges, and Applications},
JOURNAL = {Drones},
VOLUME = {9},
YEAR = {2025},
NUMBER = {6},
ARTICLE-NUMBER = {425},
}

@InProceedings{tethered_defense,
author="Durmus, Alpaslan
and Duymaz, Erol
and Baran, Mehmet",
editor="Karakoc, T. Hikmet
and Le Clainche, Soledad
and Chen, Xin
and Dalkiran, Alper
and Ercan, Ali Haydar",
title="The Use of Tethered Unmanned Aerial Vehicles in the Field of Defense and Current Developments",
booktitle="New Technologies and Developments in Unmanned Systems",
year="2023",
publisher="Springer International Publishing",
address="Cham",
%%pages="207--214",
abstract="The use of UAV systems for military reconnaissance and surveillance activities is one of the most frequently used methods. However, problems related to flight time cause restrictions in the use of UAV systems on the field of defense and security. With today's developing technologies, TUAV (Tethered Unmanned Aerial Vehicle) systems have been started to be designed to overcome the flight time limitations. TUAV systems in the field of defense are used to ensure border security, protect forward operational bases, and establish telecommunication ports. TUAV systems offer significant advantages to military bases in many aspects in reconnaissance and observation activities. They have important advantages in terms of flight time, safe flight, ease of use, autonomy, legislations, secure data transfer, and deterrence. In this study, usage examples of TUAV systems in the field of defense were examined, and examples of use in the field of defense were presented.",
}

@ARTICLE{lima_multi,
  title     = "A multi-model framework for tether-based drone localization",
  author    = "Lima, Rogerio R and Pereira, Guilherme A S",
  journal   = "J. Intell. Robot. Syst.",
  publisher = "Springer Science and Business Media LLC",
  volume    =  108,
  number    =  2,
  month     =  jun,
  year      =  2023,
  copyright = "https://www.springernature.com/gp/researchers/text-and-data-mining",
  language  = "en"
}

@Article{wo_cable_sensor,
AUTHOR = {Al-Radaideh, Amer and Sun, Liang},
TITLE = {Self-Localization of Tethered Drones without a Cable Force Sensor in GPS-Denied Environments},
JOURNAL = {Drones},
VOLUME = {5},
YEAR = {2021},
NUMBER = {4},
ARTICLE-NUMBER = {135},
}

@phdthesis{lima_thesis,
  title = {Exploiting the Advantages and Overcoming the Challenges of the Cable in a Tethered Drone System},
  school = {West Virginia University Libraries},
  author = {Rodrigues Lima,  Rogerio},
  year = {2023}
}

@inproceedings{schuster2024tactile,
  title={Tactile odometry in aerial physical interaction},
  author={Schuster, Micha and Bredenbeck, Anton and Beitelschmidt, Michael and Hamaza, Salua},
  booktitle={International Conference on Intelligent Robots and Systems (IROS)},
  %pages={8103--8110},
  year={2024},
}

@INPROCEEDINGS{indoor_uav,
  author={Xiao, Xuesu and Fan, Yiming and Dufek, Jan and Murphy, Robin},
  booktitle={2018 IEEE International Symposium on Safety, Security, and Rescue Robotics (SSRR)}, 
  title={Indoor UAV Localization Using a Tether}, 
  year={2018},
  volume={},
  number={},
  %%pages={1-6},
  keywords={Unmanned aerial vehicles;Robot sensing systems;Azimuth;Gravity;Cameras},
}

@ARTICLE{ugv_uav,
  author={Borgese, Andrea and Guastella, Dario C. and Sutera, Giuseppe and Muscato, Giovanni},
  journal={IEEE Robotics and Automation Letters}, 
  title={Tether-Based Localization for Cooperative Ground and Aerial Vehicles}, 
  year={2022},
  volume={7},
  number={3},
  %%pages={8162-8169},
  keywords={Location awareness;Sensors;Autonomous aerial vehicles;Length measurement;Robot kinematics;Global Positioning System;Azimuth;Aerial systems: applications;localization;multi-robot systems},
}

@phdthesis{tognon,
  TITLE = {{Theory and Applications for Control and Motion Planning of Aerial Robots in Physical Interaction with particular focus on Tethered Aerial Vehicles}},
  AUTHOR = {Tognon, Marco},
  NUMBER = {2018ISAT0030},
  SCHOOL = {{INSA de Toulouse}},
  YEAR = {2018},
  MONTH = Jul,
  TYPE = {Theses},
  HAL_ID = {tel-02003048},
  HAL_VERSION = {v2},
}

@book{euler1980rational,
  title={The rational mechanics of flexible or elastic bodies 1638-1788: introduction to Vol. X and XI},
  author={Euler, Leonhard},
  year={1980},
  publisher={Springer Science \& Business Media}
}

@misc{
  adam,
  title={CaAdam: Improving Adam optimizer using connection aware methods}, 
  author={Remi Genet and Hugo Inzirillo},
  year={2024},
  %eprint={2410.24216},
  archivePrefix={arXiv},
  primaryClass={cs.LG},
}

@inproceedings{elbo,
  title={Fixing a broken ELBO},
  author={Alemi, Alexander and Poole, Ben and Fischer, Ian and Dillon, Joshua and Saurous, Rif A and Murphy, Kevin},
  booktitle={International conference on machine learning},
  %pages={159--168},
  year={2018},
  %organization={PMLR}
}
